\documentclass[runningheads]{llncs}

\usepackage{eccv}

\usepackage{eccvabbrv}

\usepackage{cmap}

\usepackage{graphicx}
\usepackage{booktabs}
\usepackage{makecell}
\usepackage{tabularx}
\usepackage{threeparttable}
\usepackage{array}
\usepackage{multirow}
\usepackage{multicol}
\usepackage{adjustbox}
\usepackage[table]{xcolor}
\usepackage{graphicx}
\usepackage{caption}
\usepackage{amsmath}
\usepackage{svg}

\definecolor{bt2Handshape}{RGB}{242,95,63}      
\definecolor{bt2NonManual}{RGB}{20,156,129}     
\definecolor{bt2HandFidelity}{RGB}{47,98,189}   
\definecolor{bt2Contact}{RGB}{143,69,153}       
\definecolor{bt2Good}{RGB}{69,176,84}           
\definecolor{bt2Bad}{RGB}{214,74,65}            

\definecolor{Rcol}{rgb}{0.75,0.10,0.10}
\definecolor{Gcol}{rgb}{0.05,0.48,0.05}
\newcommand{\Dneg}[1]{{\textcolor{Rcol}{($\downarrow$#1)}}}   
\newcommand{\Dpos}[1]{{\textcolor{Gcol}{($\uparrow$#1)}}}   
\newcommand{\CDG}[1]{\cellcolor{green!28}\textbf{#1}}        
\newcommand{\CMG}[1]{\cellcolor{green!14}#1}                 
\newcommand{\CLG}[1]{\cellcolor{green!10}#1}                 
\newcommand{\CWR}[1]{\cellcolor{red!12}\textit{#1}}          
\newcommand{\CBX}[1]{{\setlength{\fboxsep}{1.5pt}\setlength{\fboxrule}{0.8pt}\fbox{#1}}}
\newcommand{\CDGb}[1]{\cellcolor{green!28}\CBX{\textbf{#1}}}
\newcommand{\CMGb}[1]{\cellcolor{green!14}\CBX{#1}}
\newcommand{\CLGb}[1]{\cellcolor{green!10}\CBX{#1}}

\usepackage[accsupp]{axessibility}  

\input glyphtounicode
\pdfglyphtounicode{endash}{2013}
\pdfglyphtounicode{emdash}{2014}
  \usepackage[breaklinks,colorlinks]{hyperref}  
  \hypersetup{
    linkcolor=eccvblue,   
    citecolor=eccvblue,   
    urlcolor=eccvblue,    
  }

\usepackage{orcidlink}

\begin{document}

\title{BackTranslation2.0 - A Linguistically Motivated Metric to Assess Sign Language Production}

\titlerunning{BT2.0}

\author{Oliver Cory\orcidlink{0009-0002-5383-7202} \and Maksym Ivashechkin\orcidlink{0000-0003-4936-1344} \and Karahan Sahin\orcidlink{0009-0003-6199-1522} \and Oline Ranum\orcidlink{0000-0001-8627-6259} \and Jianhe Low\orcidlink{0009-0009-0452-4374} \and Edward Fish\orcidlink{0009-0004-2964-8430} \and Anton Pelykh\orcidlink{0009-0005-9075-5718} \and Ozge Mercanoglu Sincan\orcidlink{0000-0001-9131-0634} \and Richard Bowden\orcidlink{0000-0003-3285-8020}}
\authorrunning{O.~Cory et al.}

\institute{Centre for Vision, Speech and Signal Processing, University of Surrey, Guildford, UK\\
\email{\{o.cory,m.ivashechkin,k.sahin,o.ranum,jianhe.low,edward.fish,a.pelykh, o.mercanoglusincan,r.bowden\}@surrey.ac.uk}\\
\url{https://cogvis-cvssp.github.io/BackTranslation2/}}

\maketitle

\label{sec:abstract}
\begin{abstract}
Sign Languages (SLs) are the primary means of communication for millions of deaf\footnote{We follow the recent convention of abandoning a distinction between Deaf and deaf and use the latter term also to refer to (deaf) members of the sign language community~\cite{kusters2017innovations,napier2016signlanguageaction}.} individuals, yet existing evaluation metrics for generated SL remain simplistic and poorly aligned with human judgements. We introduce BackTranslation2.0, a linguistically grounded evaluation metric for text-to-sign translation that moves beyond naïve backtranslation. Our approach adopts an agentic framework in which a deterministic pipeline orchestrates a suite of specialised tools to assess four scoring dimensions—grammatical correctness, phonological accuracy, motion fluency, and generation fidelity—aligned with human rater assessments.
Tool outputs are not treated independently: a set of large language model (LLM)-based cross-referential comparison modules evaluates consistency across tools and checks outputs against linguistic expectations, enabling structured reasoning over grammatical, phonological, and motion-level evidence. Final dimension scores are computed through deterministic weighted formulas over validated tool outputs.
To validate BackTranslation2.0, we introduce and evaluate on a British Sign Language (BSL) dataset rated in a human rater study across the same quality dimensions, following a protocol developed in collaboration between linguists and deaf experts, benchmarking against six baseline metrics. Our method demonstrates strong correlation with human judgements across all dimensions, providing a more comprehensive, interpretable, and linguistically principled evaluation framework for sign language production systems.

\end{abstract}

\section{Introduction}
\label{sec:intro}
Sign Languages (SLs) are rich visual–gestural languages used by millions of deaf individuals worldwide as their primary means of communication. Unlike spoken languages, SLs convey meaning through the visual modality, combining coordinated manual articulations (handshape, movement, location, and orientation) with non-manual features (NMF) such as facial expression and body posture. Computational approaches to SL processing have been studied for over two decades~\cite{starner1998real, vogler1999parallel, ong2005automatic}, with rapid progress recently across recognition~\cite{camgoz2018neural, zuo2023natural, li2025unisign, shen2025crossview}, translation~\cite{camgoz2020sign, jang2025lost, gong2024llms, yin2023gloss}, and production~\cite{stoll2020text2sign, zuo2025signs, baltatzis2024neural, yin2024t2sgpt, qi2024signgen} tasks.

Sign Language Production (SLP), the generation of SL output from spoken language text, represents a critical frontier for accessibility technology. However, a fundamental challenge remains: \textit{How do we reliably evaluate the quality of generated SL?} Existing automatic evaluation paradigms predominantly rely on backtranslation~\cite{saunders2020progressive}, where generated sign sequences are translated back into text and compared against the source sentence using metrics such as BLEU-4~\cite{papineni2002bleu}. While computationally convenient, such approaches correlate poorly with judgements from native deaf raters~\cite{saunders2020adversarial,jiang-etal-2025-meaningful}, who evaluate a far broader set of linguistic and visual properties. What's more, translation from sign to text remains unsolved for anything other than simple datasets, so using such approaches means using poorly performing recognition/translation technology to assess the performance of developing production technology. Therefore, the only reliable indicator of production or translation quality is human evaluation.

Human evaluation of SLP assesses multiple linguistic dimensions that determine whether a sequence forms a coherent and interpretable sign expression. Raters examine surface-level linguistic realisations affecting phonological and morphological components ~\cite{brentari1998prosodic, fenlon2014bsl} such as handshape accuracy, articulation location, and use of signing space and directionality, affecting naturalness and fluency of the signing, while evaluating syntactic and semantic information, such as appropriate word order and non-manual usage in relation to understandability~\cite{huenerfauth2006generating, kipp2011assessing,zhang2025toward}. Studies of human assessment therefore adopt structured, multi-faceted evaluation frameworks to capture these dimensions~\cite{pfau2012nonmanuals, Alkhazraji2021AtAD,villani2015aslpro}. However, expert annotation is expensive and time-intensive, limiting scalability~\cite{crasborn2008frequency}.

In this work, we introduce \textbf{BackTranslation2.0}, a linguistically motivated metric designed to bridge the gap between automatic evaluation and human judgement for text-to-sign translation systems. The metric evaluates four scoring dimensions aligned with five aspects rated by human evaluators: \textbf{(i)~Grammatical Correctness}, adherence to target SL syntax, including use of signing space, pronominal indexing, and directional verb agreement (directionality); \textbf{(ii)~Phonological Accuracy}, correctness of sub-lexical components, including handshape, movement, location, orientation, and non-manual features; \textbf{(iii)~Motion Fluency}, naturalness and temporal smoothness of signing motion; and \textbf{(iv)~Generation Fidelity}, visual quality and anatomical plausibility of the generated output.
We realise this multi-dimensional evaluation through an agentic framework in which a reasoning large language model (LLM) orchestrates a suite of specialised linguistic tools and structured lexical resources. Rather than treating tool outputs independently, the system performs cross-referential comparison: LLM-based modules evaluate consistency across tool outputs and assess them against linguistic expectations, enabling structured reasoning over grammatical, phonological, and motion-level evidence. Final dimension scores are computed using the weighted combination over validated tool outputs, yielding interpretable and reproducible evaluation.
Our contributions are as follows. \textbf{(1)}~We present BackTranslation2.0, a comprehensive linguistically motivated metric for evaluating text-to-sign translation that is superior to naïve backtranslation. \textbf{(2)}~We propose, to our knowledge, the first agentic framework for SL evaluation, in which specialised tools report structured evidence, the language model returns cross-referential comparison labels, and a fixed deterministic formula aggregates these into a reproducible score. \textbf{(3)}~We introduce a specialised British Sign Language (BSL) evaluation dataset rated by deaf and hearing signers across multiple quality dimensions. \textbf{(4)}~We benchmark BackTranslation2.0 against six baseline metrics, demonstrating strong correlation with human judgements across all dimensions and further generalisation at scale, providing a more interpretable, linguistically grounded paradigm for SLP.

\section{Related Works}
\label{sec:related_works}

\subsection{Automatic Evaluation of Sign Language Production}
\label{sec:rw-metrics}

SLP is the task of generating sign language output from spoken-language text.
Depending on the system, this output can take several forms: RGB video generated directly in pixel space~\cite{saunders2022signing, qi2024signgen}, articulated 3D body meshes (\eg SMPL-X~\cite{pavlakos2019expressive})~\cite{baltatzis2024neural, zuo2025signs}, pose or skeleton sequences~\cite{saunders2020progressive, tang2025sign}, or hybrid representations that combine explicit geometry with neural rendering, such as Gaussian splatting~\cite{kerbl3Dgaussians, ivashechkin2025signsplat}.
Yet automatic evaluation remains limited and inconsistent.

One family of metrics is \textit{reference-free} with respect to motion, avoiding the need for paired ground-truth signing.
The most common example is backtranslation~\cite{saunders2020progressive}, where the generated output is passed through a sign-to-text model and compared against the source sentence using text-based metrics such as BLEU-4~\cite{papineni2002bleu}.
Joint-embedding approaches such as SignCLIP~\cite{jiang-etal-2024-signclip} follow a similar principle in a pose-to-text (p2t) configuration.
Although both avoid a direct motion reference, they remain insensitive to important aspects of sign quality, including acceptable articulatory variation, grammar, spatial structure, and phonological well-formedness.

A second family is \textit{reference-based} with respect to motion, directly comparing a generated production against one or more target sequences.
In pose-based settings, Dynamic Time Warping (DTW) is commonly used to align motion trajectories under temporal mismatch and measure low-level similarity~\cite{walsh2024sign}.
Other approaches compare sequences in a learned latent space, such as the SkeletonVAE (SVAE)~\cite{skeletonvae,jiang-etal-2025-meaningful}, or in shared embedding spaces~\cite{jiang-etal-2024-signclip} in a pose-to-pose (p2p) configuration, while SiBLEU~\cite{madis} scores similarity over discretised sign representations.
Although more tolerant of surface variation than direct trajectory matching, these methods all compare against a particular target realisation and therefore cannot fully account for the grammatical, lexical, and articulatory variation present in human signing.
Modelling the distribution of acceptable signing rather than a single canonical trajectory~\cite{skeletonvae} mitigates this, but in practice requires multiple expert renditions of each sentence to estimate that distribution.
Taken together, these limitations motivate a metric that evaluates sign production across multiple complementary linguistic and visual dimensions.

Distinct from these metrics, the SLP Tier framework~\cite{BowdenRichardProf2025TaDf} is not an evaluation technique but a taxonomy, developed with the deaf community, that classifies spoken-to-sign systems by level of automation.
Analogous to the levels used for autonomous driving, it defines eight tiers, from fully manual translation (Tier~0) to fully automatic translation requiring no human intervention (Tier~7).
Its value here is in defining \emph{what matters} for a competent system through five core attributes: photo-realistic appearance, SL grammar and word order, correct use of space and direction, non-manual features, and fluid, natural motion, all required to reach the highest partially automated tier.
The taxonomy is explicitly not a substitute for accuracy and is meant to sit alongside a metric that quantifies understandability.
BackTranslation2.0 is that metric: its dimensions---grammatical correctness (with spatial and directional use), phonological accuracy (with non-manual features), motion fluency, and generation fidelity---target precisely these categories, measuring each directly and aligning evaluation with the priorities of deaf experts and the wider deaf community.

\subsection{LLMs as Evaluators and Tool-Using Agents}
\label{sec:rw-agents}

LLMs perform well not only at generation but also at structured analysis, comparison, and rubric-based evaluation~\cite{li2024llms}.
Work on LLM-as-a-judge~\cite{zheng2024judging} shows that, when guided by explicit criteria and grounded in relevant evidence, their assessments align closely with human and expert judgements, and \cite{jang2025lost} applies this property to sign-to-text translation quality.
In parallel, tool-augmented agentic frameworks extend these capabilities, allowing models to query specialised modules, compare heterogeneous outputs, retrieve external knowledge, and reason across multiple evidence sources rather than relying on a single monolithic forward pass~\cite{yao2023react, schick2024toolformer}.
Within SL specifically, SignAgent~\cite{cory2026signagent} applies agents to data annotation and curation.
SL evaluation is a natural fit for this agentic paradigm, since no single representation or model captures all relevant aspects of quality: grammaticality, phonological accuracy, motion fluency, and visual fidelity each demand different forms of evidence and analysis.
Rather than treating the LLM as the recogniser or generator itself, our framework uses it as a reasoning layer that interprets and cross-references the outputs of specialised tools together with structured linguistic knowledge.
This yields a more transparent, multi-dimensional evaluation, closer to the way human raters assess signing in practice and aligned with the SLP tiers.

\section{Method}
\label{sec:method}

We present BackTranslation2.0 (BT2), a deterministic agentic framework for linguistically grounded evaluation of text-to-sign translation.
Rather than relying on a single end-to-end metric, BT2 runs a fixed two-phase pipeline of specialised tools that assess \emph{grammatical correctness}, \emph{phonological accuracy}, \emph{motion fluency}, and \emph{generation fidelity}.  It operates in two phases:
Phase~1 extracts lexical, spatial, phonological, motion, and visual evidence from the generated signing.
Phase~2 cross-references that evidence against linguistic expectations and retrieved lexical knowledge to produce grounded dimension-level judgements.
All intermediate outputs are retained in a shared memory trace, enabling deterministic scoring and an auditable final natural-language assessment.
Full implementation details for all tools are provided in the supplementary material.

\subsection{Overview}
\begin{figure}[t]
  \centering
  \includegraphics[width=\textwidth]{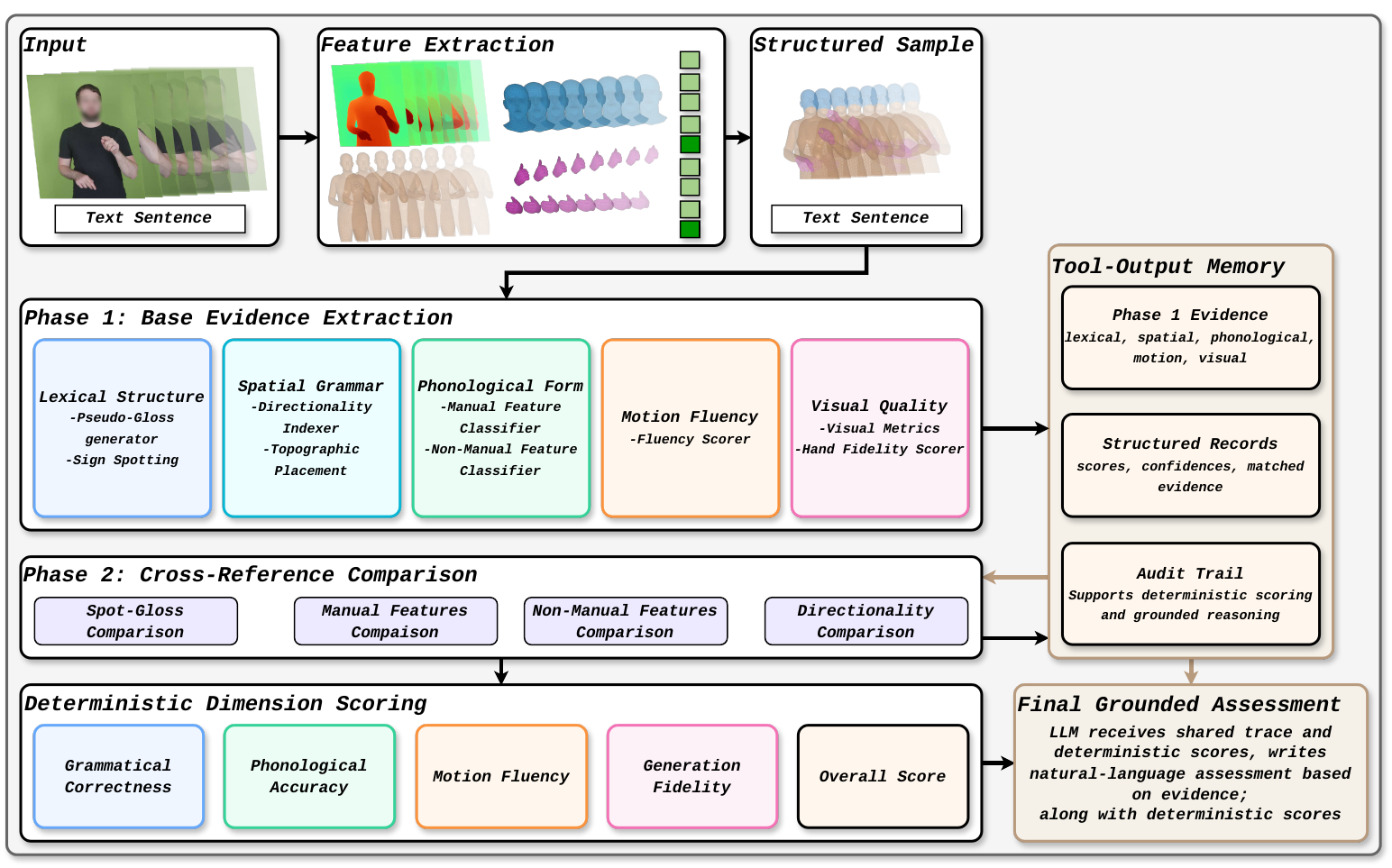}
  \caption{Overview of the BackTranslation2.0 (BT2) pipeline.
  Given a source sentence $s$ and generated sign output $\hat{y}$, multi-modal extraction produces a structured sample for segment-level and sequence-level analysis.
  Phase~1 base tools extract lexical, spatial, phonological, motion, and visual evidence, which is stored in a shared memory trace.
  Phase~2 comparison tools cross-reference this evidence against linguistic expectations to produce grounded judgements.
  Deterministic aggregation yields per-dimension scores and an overall understandability score, and a final LLM stage uses the same trace to write a grounded natural-language assessment.}
  \label{fig:architecture}
\end{figure}
Given a source sentence $s$ and a generated sign output $\hat{y}$, BT2 returns per-tool structured evidence, per-dimension scores, a deterministic overall understandability score, and a grounded natural-language assessment.
The generated sign output may be represented as RGB video, pose sequence, mesh animation, or a hybrid rendering format.
BT2 first applies multi-modal feature extraction to construct a shared structured sample,
\begin{equation}
\mathcal{X} = \Phi(s,\hat{y}) = \{x^{(1)},\dots,x^{(M)}, x^{\mathrm{seq}}\},
\end{equation}
where $x^{(1)},\dots,x^{(M)}$ denote temporal segments of the signing stream and $x^{\mathrm{seq}}$ denotes sequence-level features over the full production.
This allows some tools to operate at the \emph{segment level} (\eg sign-level lexical or phonological analysis) and others at the \emph{sequence level} (\eg fluency or global visual quality).
BT2 then executes a fixed set of $N{=}10$ base tools, followed by a fixed set of comparison tools:
\begin{equation}
\mathcal{T}_{\mathrm{base}} = \{T_1,\dots,T_N\}, \qquad
\mathcal{T}_{\mathrm{comp}} = \{C_1,\dots,C_L\}.
\end{equation}
Unlike open-ended tool-using agents, BT2 does not decide dynamically which tools to call.
Every sample follows the same deterministic pipeline, which makes the procedure reproducible, comparable across samples, and directly aligned with the target evaluation dimensions.

\noindent\textbf{Tool-output memory. }
Each tool writes a structured record into a shared memory trace,
\begin{equation}
\mathcal{M} = \{m_1,\dots,m_{N+L}\},
\end{equation}
where each memory entry stores a score, confidence, and evidence.
The comparison stage consumes this memory rather than reprocessing the raw input, and the trace is passed to the final assessment stage.
BT2 therefore provides an auditable path from low-level observations to final scores and explanation.

\subsection{Phase 1: Base Evidence Extraction}

Phase~1 extracts the evidence required for each quality dimension.
The base tools target lexical content and ordering for grammar, manual and non-manual form for phonology, motion statistics for fluency, and visual plausibility for fidelity.
In the primary extraction path, we extract SMPLX~\cite{pavlakos2019expressive} for body pose with NLF~\cite{sarandi2024neural}, MANO~\cite{romero2017mano} for hand pose with WiLoR\cite{potamias2025wilor}, FLAME~\cite{li2017flame} for face parameters with TEASER~\cite{liu2025TEASER}, RTMPose~\cite{jiang2023rtmpose} for whole-body tracking, Video-Depth-Anything~\cite{yang2024depthanything} for depth, and $\mathrm{SMPLX}_{\mathrm{depth}}$ location features (SMPL-X re-optimised with depth cues), while segmentation defines temporal units \cite{he2025hands}:
\begin{equation}
x^{(m)}=(b_m,h_m,f_m,\ell_m,r_m),
\end{equation}
where $b_m,h_m,f_m,\ell_m,r_m$ denote segment-sliced body, hand, face, location, and whole-body features, and $x^{\mathrm{seq}}$ stores their full-sequence counterparts with segment set $\mathcal{S}=\{\tau_m\}_{m=1}^{M}$.
Table~\ref{tab:base-tools} summarises the base tools, grouped below; full implementation and training details are in the supplementary material.
\begin{table}[t]
      \centering
      \caption{Base tools used in Phase~1.
      `Seg.' and `Seq.' denote segment-level and sequence-level tools. The directionality tool uses a Sign Language Graph Convolutional Network (SL-GCN)~\cite{jiang2021slgcn} indexing classifier~\cite{ranum2026s}. Use: Gram.\ = Grammatical Correctness,
  Phon.\ = Phonological Accuracy, Flu.\ = Motion Fluency,
  Fid.\ = Generation Fidelity.}
      \label{tab:base-tools}

      \scriptsize
      \setlength{\tabcolsep}{3pt}
      \renewcommand{\arraystretch}{0.9}

      \begin{tabularx}{\linewidth}{@{}>{\raggedright\arraybackslash}p{2.55cm} c >{\raggedright\arraybackslash}X >{\raggedright\arraybackslash}p{1.2cm}@{}}
          \toprule
          \textbf{Tool} & \textbf{Scope} & \textbf{Description} & \textbf{Use} \\
          \midrule
          Pseudo-Gloss & Seq. & LLM proposes BSL pseudo-gloss order. & Gram. \\
          \rowcolor{black!4} Sign Spotting & Seg. & Top-$K$ sign-embedding gloss matches per segment. & Gram. \\
          Directionality & Seg. & SL-GCN indexing/pointing classifier. & Gram. \\
          \rowcolor{black!4} Topographic Scorer & Seq. & FastDTW wrist-path direction matching. & Gram. \\
          Handshape & Seg. & Transformer handshape classifier. & Phon. \\
          \rowcolor{black!4} Location & Seg. & Location classifier from $\mathrm{SMPLX}_{\mathrm{depth}}$. & Phon. \\
          Non-Manuals & Seg. & AU-to-NMF classifier. & Phon. \\
          \rowcolor{black!4} Fluency & Seq. & Flow-matching bottleneck fluency score. & Flu. \\
          Visual Metrics & Seq. & Deterministic smoothness/spectrum score. & Flu./Fid. \\
          \rowcolor{black!4} Hand Fidelity & Seq. & Binary hand-realism logit score. & Fid. \\
          \bottomrule
      \end{tabularx}
  \end{table}

\noindent\textbf{Grammar and spatial evidence. }
Pseudo-gloss generation maps the source sentence to candidate gloss-order sequences:
\begin{equation}
\mathcal{G}=f_{\mathrm{pg}}(s),\quad
g^\star=\mathrm{First}(\mathcal{G}).
\end{equation}
Here, $f_{\mathrm{pg}}$ is the pseudo-gloss generator, $\mathcal{G}$ is its candidate sequence set, and $g^\star$ is the top returned candidate used by downstream comparison tools.
Sign spotting embeds each segment and returns top-$K$ dictionary matches:
\begin{equation}
z_m=E_{\mathrm{SR}}(\hat{y}_{\tau_m}),\quad
\hat{g}_{m,1:K}=\mathrm{TopK}_K\,\cos\!\left(z_m,E_{\mathrm{dict}}\right).
\end{equation}
Here, $\hat{y}_{\tau_m}$ is the generated signing segment on interval $\tau_m$, $E_{\mathrm{SR}}$ is the sign-embedding encoder~\cite{wong2025signrep}, $z_m$ is the segment embedding, $E_{\mathrm{dict}}$ is the lexical reference bank, $K$ is the number of returned matches, and $\hat{g}_{m,1:K}$ are per-segment candidate glosses used by the Phase~2 spot-gloss comparison.
For directionality, we combine SL-GCN~\cite{jiang2021slgcn} indexing prediction~\cite{ranum2026s} with trajectory matching:
  \begin{equation}
  \hat{c}_m=\mathbf{1}\!\left[f_{\text{SL-GCN}}(\tilde{b}_m,\tilde{h}_m)\ge\theta_{\mathrm{idx}}\right],\quad
  \hat{d}=\arg\min_{d\in\mathcal{R}}\mathrm{DTW}(w_{xz},r_d).
  \end{equation}
  Here, $\tilde{b}_m,\tilde{h}_m$ are normalised body/hand features, $\theta_{\mathrm{idx}}$ the indexing threshold, $\hat{c}_m$ the indexing decision for segment $m$, $w_{xz}$ the wrist trajectory, and $\hat{d}$ the best-matching trajectory in the reference set $\mathcal{R}$ (giving $r_d$).
  The directional match rate is then mapped to the 0--4 topographic score used downstream.

\noindent\textbf{Phonological evidence.}
Handshape and location use the same transformer architecture with spatial and temporal attention; the location branch adds contact-proximity features.
The handshape classifier predicts both hands and keeps the dominant-confidence hypothesis:
\begin{equation}
p_m^{L}=f_{\mathrm{hs}}(\nu(h_m^{L})),\quad
p_m^{R}=f_{\mathrm{hs}}(\nu(h_m^{R})),\quad
(\hat{y}_m^{\mathrm{hs}},\hat{q}_m^{\mathrm{hs}})=\arg\max_{h\in\{L,R\}}\max p_m^{h}.
\end{equation}
Here, $h_m^{L},h_m^{R}$ are left/right hand sequences, $\nu(\cdot)$ is hand-keypoint normalisation, $f_{\mathrm{hs}}$ is the handshape classifier, $p_m^{h}$ is the class posterior for hand $h$, and $(\hat{y}_m^{\mathrm{hs}},\hat{q}_m^{\mathrm{hs}})$ are the dominant handshape label and confidence.
The location classifier uses $\mathrm{SMPLX}_{\mathrm{depth}}$ features with contact-proximity cues, applies two checkpoints of the same transformer backbone, and maps mean confidence to 0--4:
\begin{equation}
u_m=\pi_{\mathrm{loc}}(\ell_m,\tau_m),\quad
s_m^{\mathrm{loc}}=4\cdot\frac{\max f_{\mathrm{loc}}^{a}(u_m)+\max f_{\mathrm{loc}}^{b}(u_m)}{2}.
\end{equation}
Here, $\ell_m$ is the segment location tensor, $\pi_{\mathrm{loc}}(\cdot)$ is the preprocessing map, $u_m$ is the model input, $f_{\mathrm{loc}}^{a},f_{\mathrm{loc}}^{b}$ are the two location classifiers, and $s_m^{\mathrm{loc}}$ is the 0--4 location score.
The non-manual classifier maps per-frame action units (AUs) to NMF categories:
\begin{equation}
\mathcal{A}_m=\{a:\mathrm{AU}_{t,a}>\eta_a,\ t\in\tau_m\},\quad
\hat{n}_m=\Gamma(\mathcal{A}_m;\mathcal{M}_{\mathrm{AU}\rightarrow\mathrm{NMF}}).
\end{equation}
Here, $\eta_a$ is the AU threshold for unit $a$, $\mathcal{A}_m$ is the active-AU set on segment $\tau_m$, $\mathcal{M}_{\mathrm{AU}\rightarrow\mathrm{NMF}}$ is the semantic mapping, $\Gamma(\cdot)$ is the segment-level aggregation map, and $\hat{n}_m$ is the inferred non-manual label set.

\noindent\textbf{Motion and fidelity evidence.}
For motion fluency, a conditional flow-matching model~\cite{low2026signspark} encodes each modality sequence, and we use the U-Net bottleneck tensor for scoring.
The bottleneck features are averaged over time and evaluated under a reference Gaussian in bottleneck space:
\begin{equation}
\begin{aligned}
v_i&=\frac{1}{T_b}\sum_{t=1}^{T_b}B_{i,:,t}, \\
\log p(v_i)&=-\frac{1}{2}\!\left[(v_i-\mu)^\top\Sigma^{-1}(v_i-\mu)+\log|\Sigma|+C\log(2\pi)\right].
\end{aligned}
\end{equation}
Here, $B_{i,:,t}$ is the U-Net bottleneck feature of sample element $i$ at bottleneck time index $t$, $T_b$ is bottleneck length, $v_i$ is the mean bottleneck vector, $(\mu,\Sigma)$ are the reference Gaussian parameters in $C$ dimensions, and $\log p(v_i)$ is the fluency evidence prior to 0--4 calibration.
The log-probability is linearly calibrated to $[0,4]$, and hand/body branches are averaged when available.
Visual metrics provide deterministic kinematic evidence from body pose:
\begin{equation}
c_{\mathrm{vis}}=w_{\mathrm{hf}}\phi_{\mathrm{hf}}+w_{\mathrm{sm}}\phi_{\mathrm{sm}}+w_{\mathrm{act}}\phi_{\mathrm{act}}+w_{\mathrm{fr}}\phi_{\mathrm{fr}},
\end{equation}
where $\phi_{\mathrm{hf}},\phi_{\mathrm{sm}},\phi_{\mathrm{act}},\phi_{\mathrm{fr}}$ encode high-frequency energy, smoothness (jerk), motion activity, and dominant-frequency plausibility, $w_{\cdot}$ are fixed weights, and $c_{\mathrm{vis}}$ is the aggregated visual-motion evidence score.
Hand fidelity trains a binary classifier on synthetic vs.\ real hand crops; the score is the mean logit over crops:
\begin{equation}
z_k=f_{\mathrm{hfid}}(I_k),\quad
s_{\mathrm{hand}}=\frac{1}{|\mathcal{K}|}\sum_{k\in\mathcal{K}} z_k.
\end{equation}
Here, $I_k$ is crop $k$, $f_{\mathrm{hfid}}$ is the hand-fidelity (real-vs-synthetic) classifier, $z_k$ is the real-vs-synthetic logit for crop $k$, $\mathcal{K}$ is the crop set, $|\mathcal{K}|$ is the number of crops, and $s_{\mathrm{hand}}$ is the mean logit (linearly mapped to 0--4 for aggregation).
Full architecture, training, and calibration settings are provided in the supplementary material.

\subsection{Phase 2: Cross-Reference Comparison}

Phase~2 maps Phase~1 outputs to grounded comparison evidence by contrasting expected and detected signals under tool-specific context.
Instead of averaging independent predictions, this stage enforces linguistically structured cross-referencing before deterministic scoring.
Phase~2 uses four fixed comparison tools: spot-gloss comparison, manual-features comparison, non-manual-features comparison, and directionality comparison.

\begin{equation}
\mathcal{E}_j=\mathcal{C}_j(a_j,b_j\mid\kappa_j),\quad
j\in\{\mathrm{sg},\mathrm{man},\mathrm{nmf},\mathrm{dir}\}.
\end{equation}
Here, $\mathrm{sg}$, $\mathrm{man}$, $\mathrm{nmf}$, and $\mathrm{dir}$ denote spot-gloss, manual-features, non-manual-features, and directionality comparison tools, respectively; $a_j$ denotes expected-side evidence, $b_j$ denotes detected-side evidence from Phase~1, $\kappa_j$ denotes tool-specific context (dictionary resources, rules, and prompt constraints), and $\mathcal{E}_j$ denotes comparison evidence written to memory.

\noindent\textbf{Tool instantiations.} The four comparison tools instantiate this general form as:
\begin{equation}
\begin{aligned}
\mathcal{E}_{\mathrm{sg}}&=\mathcal{C}_{\mathrm{sg}}\!\left(g^\star,\{\hat{g}_{m,1:K}\}_{m=1}^{M}\mid\kappa_{\mathrm{sg}}\right),\\
\mathcal{E}_{\mathrm{man}}&=\mathcal{C}_{\mathrm{man}}\!\left((g^\star,\mathcal{E}_{\mathrm{sg}}),\{(\hat{y}_m^{\mathrm{hs}},s_m^{\mathrm{loc}})\}_{m=1}^{M}\mid\kappa_{\mathrm{man}}\right),\\
\mathcal{E}_{\mathrm{nmf}}&=\mathcal{C}_{\mathrm{nmf}}\!\left((s,g^\star),\{\hat{n}_m\}_{m=1}^{M}\mid\kappa_{\mathrm{nmf}}\right),\\
\mathcal{E}_{\mathrm{dir}}&=\mathcal{C}_{\mathrm{dir}}\!\left((s,g^\star),(\{\hat{c}_m\}_{m=1}^{M},\hat{d})\mid\kappa_{\mathrm{dir}}\right).
\end{aligned}
\end{equation}
Here, $\mathcal{E}_{\mathrm{sg}}$ is spot-gloss semantic alignment evidence, $\mathcal{E}_{\mathrm{man}}$ is manual phonological compatibility evidence, $\mathcal{E}_{\mathrm{nmf}}$ is non-manual expectation-vs-detection evidence, and $\mathcal{E}_{\mathrm{dir}}$ is indexing/topographic directionality evidence.
Concretely, $\mathcal{C}_{\mathrm{sg}}$ uses deterministic prefiltering followed by LLM semantic matching on unresolved pairs, $\mathcal{C}_{\mathrm{man}}$ is dictionary-based with optional LLM fallback for low-confidence mapping, $\mathcal{C}_{\mathrm{nmf}}$ can infer expected markers via LLM or deterministic rules, and $\mathcal{C}_{\mathrm{dir}}$ is deterministic. The resulting evidence set $\mathcal{M}_{\mathrm{comp}} = \{\mathcal{E}_{\mathrm{sg}}, \mathcal{E}_{\mathrm{man}}, \mathcal{E}_{\mathrm{nmf}}, \mathcal{E}_{\mathrm{dir}}\}$ is consumed by deterministic dimension scoring.

\subsection{Deterministic Dimension Scoring}
\label{sec:aggregation}

After both phases, BT2 deterministically computes four dimension scores,
\(\mathbf{d}=[d_{\mathrm{gram}},d_{\mathrm{phon}},d_{\mathrm{flu}},d_{\mathrm{fid}}]\), each on a common 0--4 scale.
Each dimension is formed from fixed subcomponent weights, with renormalisation when a subcomponent is unavailable.
No LLM step modifies these numeric scores.

\noindent\textbf{Overall understandability. }
The authoritative overall score $u = \frac{1}{4}(d_{\mathrm{gram}} + d_{\mathrm{phon}} + d_{\mathrm{flu}} + d_{\mathrm{fid}})$ is the arithmetic mean of the four deterministic dimensions.

\subsection{Final Grounded Assessment}

The final stage receives the full memory trace \(\mathcal{M}\) together with fixed scores \(\mathbf{d}\) and \(u\).
Its role is explanation only: it produces a grounded natural-language assessment and does not recompute or replace any numeric score.
BT2 therefore returns structured per-tool evidence, fixed per-dimension scores, a fixed overall score, and an auditable grounded assessment.

\section{Experiments}
\label{sec:experiments}

We evaluate BT2 on a purpose-built BSL benchmark (Sec.~\ref{sec:dataset}) probing the categories our metric targets: grammatical correctness, phonological accuracy, motion fluency, generation fidelity, and overall understandability. We compare BT2 against reference-free and reference-based baselines using human ratings of understandability and overall quality (Sec.~\ref{sec:results}), and test whether it is sensitive to the same controlled anomalies and corruptions identified by human raters.
We then ablate each component (Sec.~\ref{sec:ablation}) and demonstrate large-scale generalisation on a synthetic corpus of procedurally corrupted sentences (Sec.~\ref{sec:scale}).

\subsection{Evaluation Dataset}
\label{sec:dataset}
Our benchmark has four complementary subsets spanning real signing, controlled corruption, synthetic production, and targeted spatial-grammar evaluation, summarised in Table~\ref{tab:dataset}.

\noindent\textbf{Human Sentence Repetition Tasks (SRT). }
This subset contains real BSL sentence-repetition recordings from six signers at different proficiency levels (three learner levels and native), each producing the same six sentences. It provides paired comparison against native reference signing and captures natural proficiency-driven quality variation.

\noindent\textbf{Known Corruptions.}
To isolate specific linguistic failure modes, we record human signers producing targeted corruptions of the same six sentences: missing non-manual features, incorrect word order, and incorrect handshape. This subset tests whether metrics penalise specific anomalies rather than only broad visual changes.

\noindent\textbf{Synthetic Productions. }
We evaluate a set of synthetic productions for the same reference content, spanning text-to-sign and re-animation systems across pose-based, mesh-based, and RGB-based production systems, including SL-specific approaches SignGAN~\cite{saunders2022signing}, SignSplat~\cite{ivashechkin2025signsplat}, SignStream~\cite{signstream2026}, and SignSparK~\cite{low2026signspark}, as well as human motion generation systems, GUAVA~\cite{zhang2025guava} and Wan2.1~\cite{wan2025wan}. This subset focuses on production-method quality under shared content.

\noindent\textbf{Directionality and topographic placement. }
We include a targeted set of sentences probing spatial grammar, including directional verbs, pronominal indexing, and topographic placement in signing space. This subset stresses phenomena that are weakly captured by conventional automatic metrics. Together, these subsets cover natural proficiency variation, controlled linguistic corruption, synthetic production quality, and spatial-grammatical correctness.

\noindent\textbf{Human study. }
We conduct a user study with 20 participants (deaf and hearing) on a benchmark subset. Participants provide 5-point Likert ratings for \emph{understandability} and \emph{overall quality}, and flag perceived anomalies (handshape, hand positioning, non-manual features, word order, visual fidelity, fluency, and jitter). This supports both human-alignment analysis and corruption-sensitivity analysis. The full protocol, developed in collaboration between linguists and deaf experts, is provided in the supplementary material.

\begin{table}[t]
    \centering
    \caption{Composition of the evaluation benchmark. Further details are provided in the supplementary material.}
    \label{tab:dataset}
    
    \scriptsize 
    \setlength{\tabcolsep}{3pt} 
    \renewcommand{\arraystretch}{0.9} 
    
    \begin{tabular}{@{} l l l @{}}
        \toprule
        \textbf{Subset} & \textbf{Content} & \textbf{Purpose} \\
        \midrule
        Human SRT & 6 signers $\times$ 6 sentences & Proficiency variation \\
        Known Corruptions & 6 sent. $\times$ 6 sign. $\times$ 3 corrupt. & Anomaly sensitivity \\
        Synthetic Productions & 6 production systems & Production-method evaluation \\
        Directionality / Topography & Targeted spatial-grammar set & Grammatical placement \\
        \bottomrule
    \end{tabular}
\end{table}

\subsection{Results}
\label{sec:results}

\begin{table}[p]
\centering
\setlength{\belowcaptionskip}{2pt}
\caption{Alignment between automatic metrics and human judgements across known corruptions, synthetics, and human SRT. `Under.' and `Qual.' denote understandability and overall quality. Metric cells show direction-normalised Pearson $r$ / Spearman $\rho$ (higher is better). \textbf{Bold} marks column-wise best values, and \textcolor{green!45!black}{green} marks BT2 cells that match or exceed the strongest reference-free baseline. DTW-MPJPE is the Mean Per-Joint Position Error after Dynamic Time Warping alignment, and SignCLIP is evaluated in pose-to-pose (p2p) and pose-to-text (p2t) configurations. The lower blocks report BT2 per-component scores and human inter-rater agreement (Gwet's AC2~\cite{gwet2008computing}, ICC~\cite{shrout1979intraclass}, within-1, Quad. weighted $\kappa$~\cite{cohen1968weighted}, and Krippendorff's $\alpha$~\cite{krippendorff2011computing} for anomaly flags).}
\label{tab:eccv-baselines-main}

\fontsize{6.3pt}{8.2pt}\selectfont
\setlength{\tabcolsep}{3pt}
\renewcommand{\arraystretch}{0.74}
\setlength{\extrarowheight}{1.1pt}
\setlength{\aboverulesep}{0.6pt}
\setlength{\belowrulesep}{1.8pt}
\setlength{\abovetopsep}{0pt}
\setlength{\belowbottomsep}{0pt}

\begin{tabular*}{\textwidth}{@{\extracolsep{\fill}} p{0.39\textwidth} r<{\,/}@{}r r<{\,/}@{}r r<{\,/}@{}r r<{\,/}@{}r @{}}
\toprule
\textbf{Method} & \multicolumn{2}{c}{\makecell[c]{\textbf{Known}\\\textbf{Corrupt.}}} & \multicolumn{4}{c}{\textbf{Synthetics}} & \multicolumn{2}{c}{\makecell[c]{\textbf{Human}\\\textbf{SRT}}} \\
\cmidrule(lr){2-3}\cmidrule(lr){4-7}\cmidrule(lr){8-9}
 & \multicolumn{2}{c}{\textbf{Under.}} & \multicolumn{2}{c}{\textbf{Under.}} & \multicolumn{2}{c}{\textbf{Qual.}} & \multicolumn{2}{c}{\textbf{Under.}} \\
\midrule
\multicolumn{9}{@{}l}{\textbf{Reference-based \,(req. source sequence)}} \\
DTW-MPJPE & 0.91 & 0.50 & -0.14 & -0.03 & -0.33 & -0.37 & 0.04 & 0.00 \\
\rowcolor{black!4} SiBLEU\cite{madis} & -0.58 & -0.50 & 0.03 & 0.31 & -0.05 & 0.31 & \textbf{0.96} & \textbf{0.90} \\
\multicolumn{9}{@{}l}{SignCLIP~\cite{jiang-etal-2024-signclip} p2p} \\
\hspace{0.75em}BSL sentence & -0.97 & -1.00 & \textbf{0.85} & 0.90 & 0.82 & 0.70 & 0.69 & 0.80 \\
\rowcolor{black!4} \hspace{0.75em}BSL segmentation & -0.55 & -0.50 & 0.75 & 0.70 & 0.63 & 0.60 & -0.46 & -0.40 \\
\hspace{0.75em}Multilingual sentence & 0.85 & \textbf{1.00} & 0.70 & 0.70 & 0.55 & 0.60 & -0.10 & -0.40 \\
\rowcolor{black!4} \hspace{0.75em}Multilingual segmentation & -0.59 & -0.50 & 0.83 & 0.70 & 0.75 & 0.60 & -0.30 & 0.00 \\
SVAE~\cite{skeletonvae} & 0.95 & \textbf{1.00} & 0.43 & 0.26 & 0.54 & 0.31 & 0.81 & 0.80 \\
\specialrule{0.12em}{0.1em}{0.18em}
\multicolumn{9}{@{}l}{\textbf{Reference-free \,(req. source text)}} \\
\multicolumn{9}{@{}l}{BT1~\cite{saunders2020progressive}} \\
\rowcolor{black!4} \hspace{0.75em}BLEU-4~\cite{papineni2002bleu} & -0.19 & 0.00 & -0.38 & -0.44 & -0.15 & -0.10 & 0.00 & 0.00 \\
\hspace{0.75em}BLEURT~\cite{sellam2020bleurt} & 0.09 & 0.50 & -0.35 & 0.03 & -0.39 & -0.09 & 0.15 & 0.35 \\
\rowcolor{black!4} \hspace{0.75em}ChrF~\cite{popovic-2015-chrf} & -0.87 & -1.00 & -0.27 & -0.09 & -0.54 & -0.37 & -0.15 & -0.35 \\
\hspace{0.75em}ROUGE~\cite{lin-2004-rouge} & -0.19 & 0.00 & -0.34 & -0.44 & -0.08 & -0.10 & 0.00 & 0.00 \\
\multicolumn{9}{@{}l}{SignCLIP~\cite{jiang-etal-2024-signclip} p2t} \\
\rowcolor{black!4} \hspace{0.75em}BSL sentence & 0.25 & 0.50 & 0.03 & 0.30 & 0.17 & 0.40 & 0.56 & 0.60 \\
\hspace{0.75em}BSL segmentation & -0.25 & -0.50 & -0.33 & -0.30 & -0.20 & -0.10 & -0.78 & -0.70 \\
\rowcolor{black!4} \hspace{0.75em}Multilingual sentence & 0.89 & 0.50 & -0.64 & -0.60 & -0.41 & -0.30 & -0.07 & -0.10 \\
\hspace{0.75em}Multilingual segmentation & 0.77 & 0.50 & -0.37 & -0.10 & -0.50 & -0.30 & -0.11 & -0.30 \\
\midrule
\rowcolor{black!4} \textbf{BT2 Overall (ours)} & \cellcolor{green!8}\textbf{1.00} & \cellcolor{green!8}\textbf{1.00} & \cellcolor{green!8}0.42 & \cellcolor{green!8}0.60 & \cellcolor{green!8}0.52 & \cellcolor{green!8}0.66 & -0.43 & -0.30 \\
\bottomrule

\addlinespace[1pt]
\toprule
\multicolumn{9}{@{}l}{\textbf{Per-component}} \\
\midrule
\textbf{BT2 Overall} & \cellcolor{green!8}\textbf{1.00} & \cellcolor{green!8}\textbf{1.00} & \cellcolor{green!8}0.42 & \cellcolor{green!8}0.60 & \cellcolor{green!8}0.52 & \cellcolor{green!8}0.66 & -0.43 & -0.30 \\
\rowcolor{black!4} BT2 Phonology & \cellcolor{green!8}0.28 & \cellcolor{green!8}0.50 & \cellcolor{green!8}0.48 & \cellcolor{green!8}0.26 & \cellcolor{green!8}0.35 & \cellcolor{green!8}-0.03 & \cellcolor{green!8}0.87 & \cellcolor{green!8}\textbf{0.90} \\
BT2 Grammar & -1.00 & -1.00 & -0.50 & -0.14 & -0.45 & -0.26 & -0.45 & -0.60 \\
\rowcolor{black!4} BT2 Fluency & \cellcolor{green!8}0.28 & \cellcolor{green!8}0.50 & \cellcolor{green!8}0.21 & \cellcolor{green!8}0.37 & \cellcolor{green!8}0.40 & \cellcolor{green!8}0.54 & -0.46 & -0.40 \\
BT2 Fidelity & \cellcolor{green!8}0.80 & \cellcolor{green!8}\textbf{1.00} & \cellcolor{green!8}0.74 & \cellcolor{green!8}\textbf{0.94} & \cellcolor{green!8}\textbf{0.83} & \cellcolor{green!8}\textbf{0.94} & -0.81 & -0.90 \\
\bottomrule

\addlinespace[1pt]
\toprule
\multicolumn{9}{@{}l}{\textbf{Human inter-rater agreement}} \\
\midrule
Gwet's AC2 & \multicolumn{2}{c}{0.69} & \multicolumn{2}{c}{0.69} & \multicolumn{2}{c}{0.68} & \multicolumn{2}{c}{0.69} \\
\rowcolor{black!4} ICC(2,1) & \multicolumn{2}{c}{-0.02} & \multicolumn{2}{c}{0.51} & \multicolumn{2}{c}{0.35} & \multicolumn{2}{c}{0.07} \\
Within-1 agreement & \multicolumn{2}{c}{0.74} & \multicolumn{2}{c}{0.79} & \multicolumn{2}{c}{0.78} & \multicolumn{2}{c}{0.75} \\
\rowcolor{black!4} Quadratic weighted $\kappa$ & \multicolumn{2}{c}{-0.03} & \multicolumn{2}{c}{0.46} & \multicolumn{2}{c}{0.35} & \multicolumn{2}{c}{0.09} \\
Handshape flag $\alpha$ & \multicolumn{2}{c}{0.24} & \multicolumn{2}{c}{0.26} & \multicolumn{2}{c}{0.26} & \multicolumn{2}{c}{0.09} \\
\rowcolor{black!4} Non-manual flag $\alpha$ & \multicolumn{2}{c}{0.27} & \multicolumn{2}{c}{0.29} & \multicolumn{2}{c}{0.29} & \multicolumn{2}{c}{-0.03} \\
Word-order flag $\alpha$ & \multicolumn{2}{c}{0.30} & \multicolumn{2}{c}{0.01} & \multicolumn{2}{c}{0.01} & \multicolumn{2}{c}{-0.01} \\
\bottomrule
\end{tabular*}

\setlength{\abovecaptionskip}{10pt}
\setlength{\belowcaptionskip}{0pt}
\caption{BT2 component ablation versus human ratings on Known Corruptions (KC), Synthetics (SC), and Human SRT. Cells report direction-normalised $r$ / $\rho$; \textcolor{Rcol}{red ($\downarrow$)} marks a drop and \textcolor{Gcol}{green ($\uparrow$)} an improvement over the full system. $^\dag$On Human SRT all ablations remain negative, indicating a structural rather than component-specific effect. The Grammar sub-score's $\rho{=}-1.00$ on KC reflects over-penalised flexible BSL word order, corrected to $+1.00$ by full aggregation.}
\label{tab:ablation}

\renewcommand{\arraystretch}{0.78}
\resizebox{\textwidth}{!}{%
\begin{tabular}{@{}l cccc@{}}
\toprule
& \textbf{KC} & \multicolumn{2}{c}{\textbf{Synthetics (SC)}} & \textbf{SRT}$^\dag$ \\
\cmidrule(lr){3-4}
\textbf{Condition} & Under.\ $r/\rho$ & Under.\ $r/\rho$ & Qual.\ $r/\rho$ & Under.\ $r/\rho$ \\
\midrule
No LLM (word heuristic)
  & $.73$\Dneg{27\%}$/.50$\Dneg{50\%}
  & $.21$\Dneg{50\%}$/.37$\Dneg{38\%}
  & $.43$\Dneg{17\%}$/.54$\Dneg{18\%}
  & $-.68$\Dneg{58\%}$/-.40$\Dneg{33\%} \\
Phase~1 only
  & $.73$\Dneg{27\%}$/.50$\Dneg{50\%}
  & $.21$\Dneg{50\%}$/.37$\Dneg{38\%}
  & $.43$\Dneg{17\%}$/.54$\Dneg{18\%}
  & $-.68$\Dneg{58\%}$/-.40$\Dneg{33\%} \\
Phase~1 + det.\ Phase~2
  & $.08$\Dneg{92\%}$/.50$\Dneg{50\%}
  & $.22$\Dneg{48\%}$/.09$\Dneg{85\%}
  & $.42$\Dneg{19\%}$/.37$\Dneg{44\%}
  & $-.63$\Dneg{47\%}$/-.50$\Dneg{67\%} \\
Full $-$ spot-gloss
  & $.73$\Dneg{27\%}$/.50$\Dneg{50\%}
  & $.31$\Dneg{26\%}$/.37$\Dneg{38\%}
  & $.41$\Dneg{21\%}$/.49$\Dneg{26\%}
  & $-.68$\Dneg{58\%}$/-.40$\Dneg{33\%} \\
Full $-$ manual feats.
  & $.56$\Dneg{44\%}$/.50$\Dneg{50\%}
  & $.26$\Dneg{38\%}$/.37$\Dneg{38\%}
  & $.34$\Dneg{35\%}$/.49$\Dneg{26\%}
  & $-.66$\Dneg{53\%}$/-.70$\Dneg{133\%} \\
Full $-$ non-manual
  & $.35$\Dneg{65\%}$/.50$\Dneg{50\%}
  & $.50$\Dpos{19\%}$/.54$\Dneg{10\%}
  & $.54$\Dpos{4\%}$/.60$\Dneg{9\%}
  & $-.09$\Dpos{79\%}$/-.10$\Dpos{67\%} \\
Full $-$ topo.\ dir.
  & $.88$\Dneg{12\%}$/.50$\Dneg{50\%}
  & $.46$\Dpos{10\%}$/.43$\Dneg{28\%}
  & $.52$\Dpos{0\%}$/.60$\Dneg{9\%}
  & $-.34$\Dpos{21\%}$/-.10$\Dpos{67\%} \\
\midrule
\rowcolor{green!8}
\textbf{BT2 (ours)}
  & $\mathbf{1.00/1.00}$
  & $\mathbf{.42/.60}$
  & $\mathbf{.52/.66}$
  & $\mathbf{-.43/-.30}$ \\
\bottomrule
\end{tabular}}
\end{table}

Table~\ref{tab:eccv-baselines-main} compares BT2 with reference-free and reference-based baselines using direction-normalised Pearson $r$ and Spearman $\rho$ agreement with human labels.
Pearson $r$ captures linear agreement in score magnitude, while Spearman $\rho$ captures rank-order agreement.
Direction normalisation is applied so higher values always indicate stronger alignment with human judgements, including for distance-style baselines.
The same test cases can therefore show different $r$ and $\rho$ values when magnitude spacing and rank ordering disagree.
Reference-free methods rely on source-sentence or back-translated text signals, while reference-based methods compare against a target signing realisation in motion or latent space.
BT2 combines category-level linguistic and visual evidence into deterministic scores, and uses the final language model to provide explanation text.

Against reference-free baselines, BT2 Overall is strongest in three of the four columns: \textit{Known Corruptions} and both synthetic columns.
\textit{Human SRT} is the only column where BT2 Overall trails reference-free baselines.
On \textit{Known Corruptions}, BT2 Overall reaches `1.00 / 1.00' and matches or exceeds the strongest reference-based alternatives.

The per-component block shows that BT2 remains competitive with reference-based methods in key settings.
BT2 Phonology reaches `0.87 / 0.90' on \textit{Human SRT}, matching the best reference-based Spearman and clearly exceeding the strongest reference-free result.
In this real SRT subset, sentence content is fixed and capture conditions are matched, so raters can separate signer proficiency primarily through manual and non-manual form differences, which aligns with the strong phonology result.
BT2 Fidelity is strongest on synthetic quality (`0.83 / 0.94') and remains above both reference-free baselines and the strongest reference-based alternative on that synthetic-quality column.
By contrast, BT2 Overall (`-0.43 / -0.30') and BT2 Fidelity (`-0.81 / -0.90') are weaker in \textit{Human SRT}, suggesting that global-quality judgements in this subset are influenced by factors beyond signer-linguistic skill.
BT2 Grammar is also weaker in \textit{Human SRT} (`-0.45 / -0.60') because natural sign production allows more flexible word placement than tightly controlled corruption settings.

\textit{Human SRT} is intrinsically difficult for rank-based evaluation because of score compression and low inter-rater agreement: raters exhibit a ceiling effect (74\% of ratings ${\geq}4$), signer means span only a 0.55-point window on the 1--5 scale, and ICC${\approx}0$.
BT2's phonological signal ($\rho{=}0.90$) succeeds where genuine variation exists, while the negative Overall most likely reflects the difficulty of reliable ranking in this highly compressed regime.
Notably, \textit{Known Corruptions}---also human-signed, but with a signer \emph{deliberately} introducing handshape, location, and word-order errors---reaches $\rho{=}1.00$, confirming that BT2 responds to signing quality rather than signing origin.

The inter-rater block clarifies why grammatical alignment is harder to capture.
Word-order flag agreement is highest in \textit{Known Corruptions} ($\alpha=0.30$), but near zero in \textit{Synthetics} ($\alpha=0.01$) and \textit{Human SRT} ($\alpha=-0.01$), where word-order variation is less overt and consistently labelled.
This pattern is consistent with greater word-placement flexibility in natural signing, and helps explain why grammar-focused correlations are weaker than phonological ones.

Figure~\ref{fig:eccv-anomaly-alignment} measures anomaly alignment directly.
BT2 Final shows the strongest score (`0.66'), ahead of SignCLIP p2p (`0.32') and SiBLEU (`0.27'), and Panel B shows the expected negative trend between BT2 score and anomaly rate.

\subsection{Ablation}
\label{sec:ablation}

We ablate seven variants of BT2 against human ratings to isolate the contribution of each component and of the language model itself (Table~\ref{tab:ablation}).
Each variant either removes a base or comparison tool or replaces the Phase~2 language model with a deterministic fallback; full definitions are provided in the supplementary material.
Every ablation reduces the Known-Corruptions Spearman $\rho$ from 1.00.
Replacing the Phase~2 language model with deterministic fallbacks reduces synthetic-understandability $\rho$ by 85\% (from 0.60 to 0.09), showing the gain comes from reasoning, not engineering alone.
Removing it entirely (No LLM) leaves only Phase~1, since both calls are essential to comparison.
Crucially, the negative \textit{Human SRT} Overall correlation persists across all seven ablations, with no variant winning on any category; this confirms the negative result is structural to the compressed-rating subset rather than a removable component.
The per-component sub-scores (lower block of Table~\ref{tab:eccv-baselines-main}) are complementary rather than redundant: Fidelity reaches $\rho{=}0.94$ on \textit{Synthetics}, while Phonology reaches $\rho{=}0.90$ on \textit{Human SRT}, where proficiency shows through articulatory form.

\begin{table}[t]
\centering
\caption{Per-dimension win rates on synthetic BSL Corpus corruptions (uncorrupted BT2 ${\geq}2.5$, $n{=}180$), conditioned on which classifier fires. \colorbox{green!28}{\textbf{Dark}}\,${\geq}80\%$; \colorbox{green!14}{mid}\,65--79\%; \colorbox{green!10}{light}\,55--64\%; \colorbox{red!12}{\textit{red}}\,wrong-way. \textbf{Bordered} cells mark the overall score and matched target dimensions; columns are the measured categories (overall, phonology, handshape/location accuracy, grammar, fluency, fidelity) and rows the corruption type \emph{(left)}.}
\label{tab:fullmatrix}
\setlength{\tabcolsep}{3pt}
\renewcommand{\arraystretch}{0.95}
\begin{minipage}[t]{0.56\textwidth}
\centering
\vspace{0pt}
\scriptsize
\resizebox{\linewidth}{!}{%
\begin{tabular}{@{}l r ccccccc@{}}
\toprule
\textbf{Corr.} & $n$ & \textbf{Ovrl} & \textbf{Phon.} & \textbf{HS} & \textbf{Loc} & \textbf{Gram} & \textbf{Flu.} & \textbf{Fid.} \\
\midrule
\multicolumn{9}{@{}l}{\textit{Handshape classifier fires:}}\\
Handshape  & 138 & \CMGb{76\%} & \CMG{73\%} & \CDGb{86\%} & \CMG{75\%} & \CLG{60\%} & \CWR{28\%} & \CMG{65\%} \\
Location   & 129 & \CMGb{75\%} & \CMG{76\%} & \CDG{82\%} & \CMGb{74\%} & \CLG{61\%} & 52\% & \CLG{59\%} \\
Word-order & 134 & \CMGb{75\%} & \CMG{73\%} & \CDG{82\%} & \CMG{73\%} & \CLGb{58\%} & 43\% & \CLG{58\%} \\
Temporal   & 121 & \CMGb{79\%} & \CMG{71\%} & \CMG{79\%} & \CMG{75\%} & \CLG{60\%} & \CLGb{55\%} & \CMG{66\%} \\
Fidelity   & 137 & \CMGb{74\%} & \CMG{73\%} & \CDG{85\%} & \CMG{73\%} & \CLG{60\%} & \CWR{26\%} & \CLGb{64\%} \\
\midrule
\multicolumn{9}{@{}l}{\textit{Location classifier fires:}}\\
Handshape  &  87 & \CMGb{69\%} & \CMG{68\%} & \CDGb{87\%} & \CDG{98\%} & \CLG{57\%} & \CWR{22\%} & \CLG{62\%} \\
Location   &  93 & \CMGb{68\%} & \CMG{72\%} & \CDG{83\%} & \CDGb{95\%} & \CLG{61\%} & 51\% & \CLG{62\%} \\
Word-order &  88 & \CMGb{78\%} & \CMG{73\%} & \CDG{88\%} & \CDG{94\%} & \CLGb{56\%} & 47\% & \CLG{64\%} \\
Temporal   &  78 & \CMGb{78\%} & \CMG{67\%} & \CDG{87\%} & \CDG{96\%} & \CLG{56\%} & \CBX{51\%} & \CMG{69\%} \\
Fidelity   &  88 & \CMGb{65\%} & \CLG{64\%} & \CDG{85\%} & \CDG{97\%} & \CLG{59\%} & \CWR{27\%} & \CMGb{66\%} \\
\bottomrule
\end{tabular}}
\end{minipage}\hfill
\begin{minipage}[t]{0.40\textwidth}
\centering
\vspace{0pt}
\includegraphics[width=0.78\linewidth]{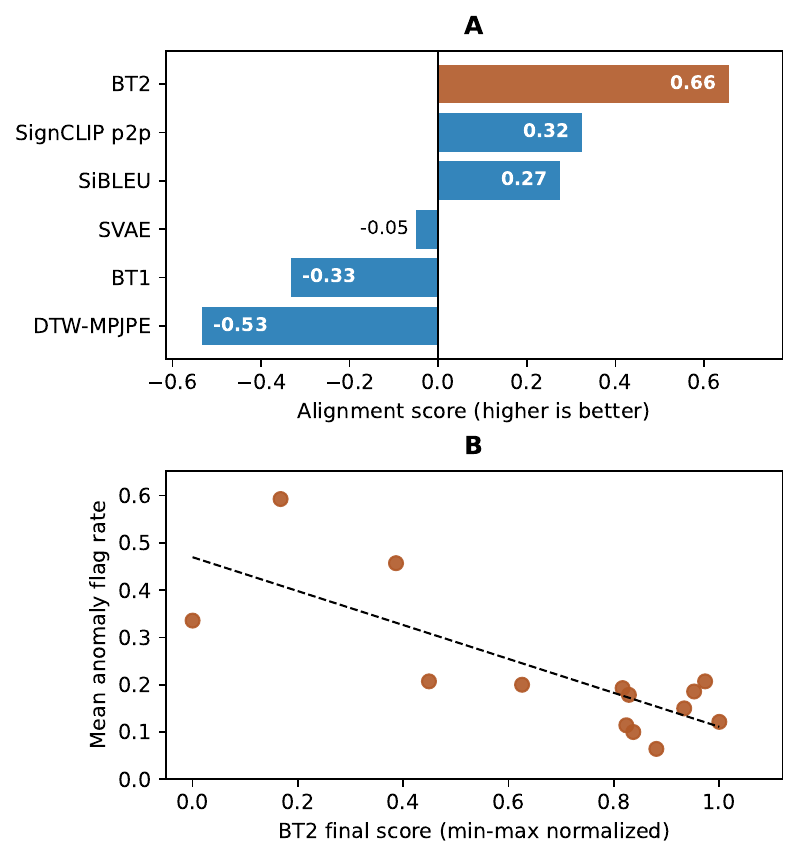}
\end{minipage}

\setlength{\belowcaptionskip}{-2pt}
\captionof{figure}{Anomaly-alignment summary for the primary comparison set. Panel A: higher values indicate stronger penalisation of videos with more flagged anomalies. Panel B: BT2 final score against total anomaly rate across the 14 active questions \emph{(right)}.}
\label{fig:eccv-anomaly-alignment}
\end{table}

\begin{table}[tb]
    \centering
    \caption{Sentence-level directional predictions across clause pairs. \textbf{Bold} words are directional verbs; \colorbox{green!8}{green pronouns} are extracted grammatical-person entities; arrows ($1\rightarrow2$) denote predicted agreement direction between person loci (indices 1/2/3 = first/second/third person). NDV: no directional verb.}
    \label{tab:surrey_directionality_examples}

    \scriptsize
    \renewcommand{\arraystretch}{1.2}

    \begin{tabular}{@{}>{\raggedright\arraybackslash}p{8.1cm}@{\hspace{0.6cm}}l@{\hspace{4pt}}c@{}}
        \toprule
        \textbf{Sentence} & \textbf{Predicted} & \textbf{Matches GT} \\
        \midrule
        \colorbox{green!8}{I} will \textbf{help} \colorbox{green!8}{you} learn some signs but \colorbox{green!8}{you} can \textbf{ask} \colorbox{green!8}{him} & $1\rightarrow2$, $2\rightarrow3$ & $\checkmark$ \\

        \rowcolor{gray!10} \colorbox{green!8}{I} will \textbf{help} \colorbox{green!8}{you} learn some signs but \colorbox{green!8}{I} can \textbf{ask} \colorbox{green!8}{him} & $1\rightarrow2$, $1\rightarrow3$ & $\checkmark$ \\

        \colorbox{green!8}{I} will \textbf{help} \colorbox{green!8}{him} learn some signs but \colorbox{green!8}{you} can \textbf{ask} \colorbox{green!8}{him}. & $1\rightarrow3$, NDV & $\checkmark$ \\

        \rowcolor{gray!10} \colorbox{green!8}{She} will \textbf{help} \colorbox{green!8}{you} learn some signs but \colorbox{green!8}{he} can \textbf{ask} \colorbox{green!8}{me}. & NDV, $1\rightarrow3$ & $\checkmark$ \\

        \colorbox{green!8}{I} will \textbf{help} \colorbox{green!8}{you} learn some signs but \colorbox{green!8}{you} can \textbf{ask} \colorbox{green!8}{him} & NDV, NDV & $\checkmark$ \\
        \bottomrule
    \end{tabular}
\end{table}

\subsection{Corruption Sensitivity and Generalisation}
\label{sec:scale}

Beyond the human-rated benchmark, we test corruption sensitivity at scale on a separate synthetic corpus: we generate BSL Corpus~\cite{schembri2013building} sentences from continuous segments and corrupt them across five modality types (handshape, location, word order, temporal, and fidelity), retaining 900 matched pairs ($n{=}180$) scoring ${\geq}2.5/4$ under BT2; details are in the supplementary material.
Table~\ref{tab:fullmatrix} reports win rates conditioned on which classifier fires.
Across all five corruption types, BT2 ranks the uncorrupted sequence above its corruption on the overall score (65--79\%) and on every matched target dimension: handshape (79--88\%), location (73--98\%), and grammar (56--61\%).
Fluency is deliberately wrong-way for the handshape and fidelity corruptions, which alter phonological form without disrupting motion, confirming BT2 localises each degradation to its target dimension.
Where no difference is detected, BT2 abstains rather than penalising.

\begin{figure}[t]
    \centering
    \includegraphics[width=1.0\linewidth]{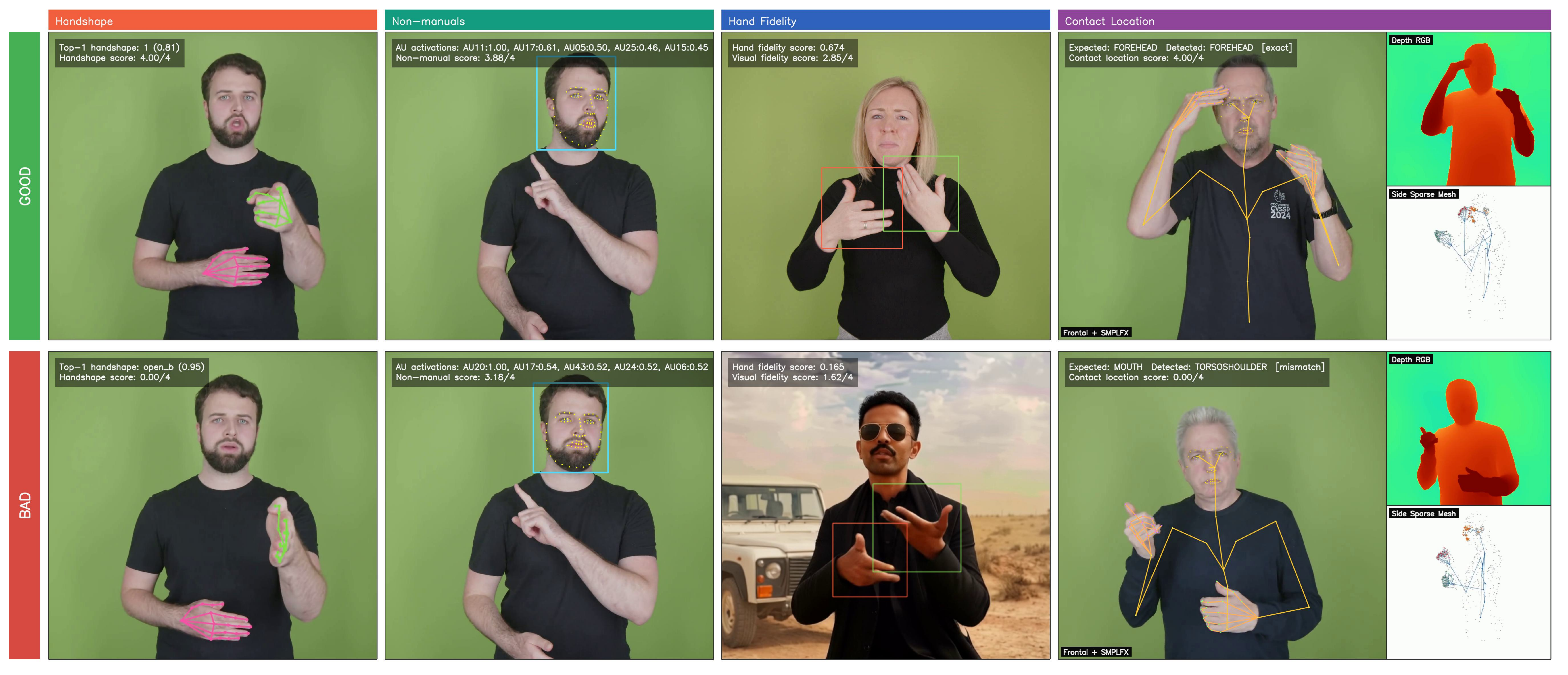}
      \caption{
  Qualitative evaluation of BT2 visual-base tools; overlays show predicted class and score, with the \textcolor{bt2Good}{top row} rewarding good production and the \textcolor{bt2Bad}{bottom row} penalising poor production.
  Columns (left to right): \textcolor{bt2Handshape}{Handshape}, \textcolor{bt2NonManual}{Non-manuals}, \textcolor{bt2HandFidelity}{Hand fidelity}, \textcolor{bt2Contact}{Contact location}.
  Handshape and Non-manuals compare best production against matched corruptions; Hand fidelity contrasts a reference clip with poor-handed synthesis; Contact location shows correct (top) versus missing (bottom) contact via a frontal SMPL-X overlay, depth map, and side sparse view.}
  \label{fig:eccv-qualitative-tools}
\end{figure}

\subsection{Qualitative Analysis}
\label{sec:qualitative}

Figure~\ref{fig:eccv-qualitative-tools} illustrates per-tool behaviour: the Handshape and Contact-location columns clearly separate good from corrupted production, the Non-manuals column reflects the presence or absence of expected facial markers, and the hand-fidelity panel separates a reference clip from degraded synthesis, supporting the per-tool classification behind the quantitative results.
Table~\ref{tab:surrey_directionality_examples} provides sentence-level directional-verb evidence for individual comparison tools.
Across the shown clause pairs, the extracted person-locus sequences match the intended directional patterns, illustrating that the directional stack captures clause-conditioned role transfer rather than only surface motion similarity.
Further qualitative material, including the motion-fluency visualisation and a full reasoning trace, is provided in the supplementary material.

\section{Conclusion}
\label{sec:conclusion}
We introduced BackTranslation2.0 (BT2), a linguistically grounded metric for SLP that decomposes assessment into interpretable dimensions aligned with human rater criteria. It aligns more closely with human judgements than existing reference-free and reference-based baselines, gives substantially richer diagnostic insight, and captures phenomena conventional metrics miss---spatial grammar, directionality, and phonological mismatches. Beyond correlation gains, it provides interpretable evidence at multiple levels, from individual tool outputs to category-level scores, supporting not only benchmarking but also the analysis of failure modes in production systems. By grounding evaluation in structured visual and linguistic evidence rather than text surrogates or surface motion similarity, it offers a more faithful and practically useful framework for SLP assessment.

\section*{Acknowledgements}
{\sloppy
This work was supported by EPSRC grant APP24554 (SignGPT-EP/Z535370/1), EPSRC grant APP78083 (UMCS UKRI3927) and through funding from Google.org via the AI for Global Goals scheme.
The authors acknowledge the use of Isambard-AI National AI Research Resource (AIRR) funded by UK DSIT via UKRI and STFC [ST/AIRR/I-A-I/1023].
This work reflects only the authors' views and the funders are not responsible for any use that may be made of the information it contains.\par}

\bibliographystyle{splncs04}
\bibliography{main.bib}

\end{document}